\newtheorem{theorem}{Theorem}
\newtheorem{lemma}{Lemma}

\documentclass[12pt]{article}

\def\tr{\mathop{\rm tr}}
\def\sgn{\mathop{\rm sgn}}
\def\sign{\mathop{\rm sign}}
\def\R{\mathop{\mathbb R}}
\def\bx{\mathop{\bf x}}
\def\bX{\mathop{\bf X}}
\def\by{\mathop{\bf y}}
\def\bY{\mathop{\bf Y}}
\def\bz{\mathop{\bf z}}
\def\bZ{\mathop{\bf Z}}
\def\bs{\mathop{\bf s}}
\def\bB{\mathop{\bf B}}
\def\be{\mathop{\bf e}}
\def\bI{\mathop{\bf I}}
\def\bH{\mathop{\bf H}}
\def\bM{\mathop{\bf M}}
\def\bxi{\mathop{\bm \xi}}
\def\bepsilon{\mathop{\bm \epsilon}}
\def\bOmega{\mathop{\bm \Omega}}
\def\bSigma{\mathop{\bm \Sigma}}
\def\bGamma{\mathop{\bm \Gamma}}
\def\bDelta{\mathop{\bm \Delta}}
\def\cA{\mathop{\cal A}}
\def\Dist{\mathop{\rm Dist}}
\def\Sen{\mathop{\rm Sen}}
\def\Spe{\mathop{\rm Spe}}
\def\Mcc{\mathop{\rm Mcc}}
\def\TP{\mathop{\rm TP}}
\def\TN{\mathop{\rm TN}}
\def\FP{\mathop{\rm FP}}
\def\FN{\mathop{\rm FN}}
\usepackage{graphics,latexsym,amssymb,epsfig,bm,fullpage,setspace,amsmath,textcomp,url,xcolor,enumerate}
\usepackage[pagewise,right]{lineno}

\begin{document}

\pagenumbering{arabic}

\setcounter{page}{1}

\title{\Large \bf Joint estimation of sparse multivariate regression and conditional graphical models}

\author{
    Junhui Wang\\
    Department of Mathematics, Statistics,\\
    and Computer Science\\
    University of Illinois at Chicago\\
    Chicago, IL 60607\\
}

\date{}
\maketitle

\begin{abstract}
Multivariate regression model is a natural generalization of the classical univariate regression model for fitting multiple responses. In this paper, we propose a high-dimensional multivariate conditional regression model for constructing sparse estimates of the multivariate regression coefficient matrix that accounts for the dependency structure among the multiple responses. The proposed method decomposes the multivariate regression problem into a series of penalized conditional log-likelihood of each response conditioned on the covariates and other responses. It allows simultaneous estimation of the sparse regression coefficient matrix and the sparse inverse covariance matrix. The asymptotic selection consistency and normality are established for the diverging dimension of the covariates and number of responses. The effectiveness of the proposed method is also demonstrated in a variety of simulated examples as well as an application to the Glioblastoma multiforme cancer data.
\end{abstract}


\doublespace

\section{Introduction} 

Multivariate regression model is a key statistical tool for analyzing dataset with multiple responses. A standard approach is to decompose the multivariate regression model and fit each response via a marginal univariate regression model. However, this approach is suboptimal in general as it completely ignores the dependency structure among the responses. For example, the gene expressions of many genes are strongly correlated due to the shared genetic variants or other unmeasured common regulators (Kendziorski et al., 2006). With the dependency structure appropriately incorporated, one would naturally expect a more efficient multivariate regression model in terms of both estimation and prediction. Furthermore, the dependency structure among the responses can be nicely interpreted in a graphical model under the multivariate Gaussian assumption (Edwards, 2000), where two Gaussian responses are connected in the graph if the corresponding entry in the precision matrix (inverse covariance matrix) is nonzero.

In literature, to model the multivariate regression problem, Breiman and Friedman (1997) proposed the curd and whey method to improve the prediction performance by utilizing the dependency among responses. The curd part fits a univariate regression model for each response against the covariates, and the whey part refits each response against the fitted values from the curd part. However, the method is developed in the low dimensional setup, and does not address the challenges when the data dimension is diverging. Yuan et al. (2007) and Chen and Huang (2012) proposed the high dimensional reduced-rank regression model, which assumes that all marginal regression functions reside in a common low dimensional space. This approach focuses on dimension reduction and largely replies on the reduced-rank assumption. Turlach et al. (2005) imposed the sparsity in the regression model through a $L_{\infty}$-norm penalty of the coefficient matrix. This method is able to identify sparsity, but may produce bias for model estimation due to the $L_{\infty}$-norm penalty. The recent work by Rothman et al. (2010), Yin and Li (2011) and Lee and Liu (2012) formulated the multivariate regression problem in a penalized log-likelihood framework, so that it allows joint estimation of the multivariate regression model and the conditional Gaussian graphical model. This formulation requires an alternating optimization scheme, which is computationally expensive and can not guarantee global optimum. 

In this paper, we propose a multivariate conditional regression model to tackle the multivariate regression problem with diverging dimension. The key idea is to formulate the protblem as the conditional log-likelihood function of each response conditioned on the covariates and other responses. The conditional log-likelihood function is then equipped with the adaptive Lasso penalty (Zou, 2006) to facilitate joint estimation of the sparse multivariate regression coefficient matrix and the sparse precision matrix. The proposed model leads to a series of augmented adaptive Lasso regression models, which can be efficiently solved by any existing optimization package. More importantly, its asymptotic properties are established in terms of the estimation consistency and selection consistency with diverging dimension. In specific, the dimension of covariates and the number of responses are allowed to diverge in an exponential order of the sample size. Numerical experiments with both simulated and real examples also support the effectiveness of the proposed method. 

The rest of the paper is organized as follows. Section 2 provides a brief introduction to the multivariate regression model, with an emphasis on the penalized log-likelihood method. Section 3 describes the proposed penalized conditional log-likelihood method in details, with theoretical justification in Section 4 and numerical experiments in Section 5. Section 6 contains a discussion, and the Appendix is devoted to the technical proofs.

\section{Prelimilaries}

In a multivariate regression setting, supposed that the training dataset consists of $({\bx}_i,{\by}_i)_{i=1}^n$, where ${\bx}_i=(x_{i1},\ldots,x_{ip})^T \in {\R}^p$ and ${\by}_i=(y_{i1},\ldots,y_{iq})^T \in {\R}^q$. Let $\bX=({\bx}_1,\ldots,{\bx}_n)^T$ and $\bY=({\by}_1,\ldots,{\by}_n)^T$ be the $n \times p$ design matrix and $n \times q$ response matrix, and let ${\bx}^j=(x_{1j},\ldots,x_{nj})^T$ and ${\by}^k=(y_{1k},\ldots,y_{nk})^T$ be the $j$-th covariate and the $k$-th response. For simplicity, the covariates and responses are centered, so that
$$
\sum_{i=1}^n x_{ij} = 0,~\sum_{i=1}^n y_{ik} = 0;~j=1,\ldots,p;~k=1,\ldots,q,
$$
A standard multivariate regression model is then formulated as
\begin{equation}
\bY = \bX \bB + \be,
\label{eqn:multivar}
\end{equation}
where $\bB=(\beta_1,\ldots,\beta_q)$ with $\beta_k=(\beta_{1k},\ldots,\beta_{pk})^T \in {\R}^p$ being the regression coefficient for the $k$-th response, and $\be=(e_1,\ldots,e_n)^T$ with $e_i=(e_{i1},\ldots,e_{iq})^T \in {\R}^q$ being the $i$-th error vector. The random vector $e_i$'s are assumed to be independent and identically sampled from a $q$-dimensional Gaussian distribution $N_q(0,\bSigma)$ with positive definite $\bSigma=(\sigma_{st})_{s,t=1}^q$. 

The maximum likelihood formulation of (\ref{eqn:multivar}), after dropping constant terms, yields that
\begin{equation}
\min_{\bB, \bOmega}~-\log |\bOmega| + \tr \Big( (\bY - \bX \bB) \bOmega (\bY - \bX \bB)^T \Big),
\label{eqn:multi_ls}
\end{equation}
where $\bOmega = \bSigma^{-1}=(\omega_{st})_{s,t=1}^q$ is also positive definite and known as the precision matrix. The precision matrix is closely connected with the Gaussian graphical models (Edward, 2000) since the conditional dependency structure among the responses can be fully determined by $\bOmega$. Specifically, $\omega_{st}=0$ implies that the $s$-th and $t$-th responses are conditionally independent given the covariates and other response variables. 

When the dimension of covariates is large, it is generally believed that the responses only rely on a small proportion of them, while other covariates are noise and provide no information about the responses at all. In addition, when the number of responses is large, the dependency structure among responses becomes sparse as some responses may have little relationship with each other. Therefore, penalized log-likelihood approach has been widely employed to analyze the multivariate regression model in literature, including Rothman et al. (2010), Yin and Li (2011) and Lee and Liu (2012). The penalized likelihood approach can be formulated as
\begin{equation}
\min_{\bB, \bOmega}~-\log |\bOmega| + \tr \Big( (\bY - \bX \bB) \bOmega (\bY - \bX \bB)^T \Big) + \lambda_{1n} p_1(\bB) + \lambda_{2n} p_2(\bOmega),
\label{eqn:multi_pls}
\end{equation}
where $p_1(\bB)$ and $p_2(\bOmega)$ are sparsity-encouraging penalties, such as the adaptive Lasso penalties $p_1(\bB)=\sum_{j,k} u_{jk} |\beta_{jk}|$ and $p_2(\bOmega)=\sum_{s \neq t} v_{st} |\omega_{st}|$ with weights $u_{jk}$ and $v_{st}$, and $\lambda_{1n}$ and $\lambda_{2n}$ are two tuning parameters. To optimize (\ref{eqn:multi_pls}), alternative updating scheme is used. It updates $\bB$ and $\bOmega$ separately pretending the other party is fixed. In specific, when $\bB$ is fixed, (\ref{eqn:multi_pls}) can be solved via the graphical Lasso algorithm (Friedman, 2008), and when $\bOmega$ is fixed, (\ref{eqn:multi_pls}) can be solved via the coordinate descent algorithm (Lee and Liu, 2012). However, as pointed out in Yin and Li (2011) and Lee and Liu (2012), the alternative updating scheme can not guarantee the global optimum, and is often computationally expensive and thus not practically scalable.


\section{Proposed Methodology}

In this section, a new estimation method based on penalized conditional log-likelihood is developed for jointly estimating the sparse multivariate regression coefficient matrix and the sparse precision matrix. The key idea is motivated from the simple fact that given the model ${\by}|{\bx} \sim N_q ({\bB}^T {\bx}, \bSigma)$ in (\ref{eqn:multivar}), 
\begin{equation}
{\by}^k | ({\bX}, {\bY}^{-k}) \sim N_n({\bX} \beta_k + ({\bY}^{-k} - {\bX} {\bB}_{-k}) \gamma_k, \tilde{\sigma}_{kk} {\bI}_n),
\label{eqn:cond_dist}
\end{equation}
for any $k=1,\ldots,q$, where ${\bY}^{-k}$ denotes the response matrix without ${\by}^k$, ${\bB}_{-k}$ denotes the coefficient matrix without $\beta_k$, $\tilde{\sigma}_{kk} = {\sigma}_{kk} - {\bSigma}_{-k,k}^T {\bSigma}_{-k,-k}^{-1} {\bSigma}_{-k,k}$, $\beta_k$ stays the same as in (\ref{eqn:multivar}), and 
\begin{equation}
\gamma_k = {\bSigma}_{-k,-k}^{-1} {\bSigma}_{-k,k} = -\frac{\bOmega_{-k,k}}{\omega_{kk}}.
\label{eqn:spar_Omega}
\end{equation}
Since $\omega_{kk}$ is always positive, it follows from (\ref{eqn:spar_Omega}) that $-\sgn(\gamma_k)=\sgn(\bOmega_{-k,k})$, where $\sgn(\gamma_k)=(\sign(\gamma_{1k}),\ldots,\sign(\gamma_{k-1,k}),\sign(\gamma_{k+1,k}),\ldots,\sign(\gamma_{q,k}))^T$ with $\sign(0)=0$ for convenience. Consequently, the sparsity in $\bOmega$ can be determined by whether $\gamma_{sk}=0$ or not, and the sparsity in $\bB$ can be determined by whether $\beta_{jk}=0$ or not.

To allow joint estimation of the sparse multivariate regression coefficient matrix and the sparse precision matrix, we then formulate the model in (\ref{eqn:cond_dist}) as a series of penalized conditional regressions of each response against the covariates and other responses. In specific, for the $k$-th response, 
\begin{equation}
\min_{\beta_k,\gamma_k}~\|{\by}^{k} - {\bX} \beta_k - ({\bY}^{-k}-\bX {\bB}_{-k}) \gamma_k\|_2^2 + \lambda_{1n} p_1(\beta_k) + \lambda_{2n} p_2(\gamma_k),
\label{eqn:cond_pls}
\end{equation}
where $\|\cdot\|_2$ is the usual Euclidean norm, $p_1(\beta_k) = \sum_{j=1}^p u_{jk} |\beta_{jk}|$ and $p_2(\gamma_k)=\sum_{s \neq k} v_{sk} |\gamma_{sk}|$ are the adaptive Lasso penalties. When $\bB_{-k}$ in (\ref{eqn:cond_pls}) is replaced by an initial consistent estimate $\widehat{\bB}^{(0)}_{-k}$, the final formulation for the proposed multivariate conditional regression model is  
\begin{equation}
\min_{\bB,\bGamma}~\sum_{k=1}^q \|{\by}^{k} - {\bX} \beta_k - ({\bY}^{-k}-\bX \widehat{\bB}^{(0)}_{-k}) \gamma_k\|_2^2 + \lambda_{1n} \sum_{k=1}^q p_1(\beta_k) + \lambda_{2n} \sum_{k=1}^q p_2(\gamma_k).
\label{eqn:cond_pls0}
\end{equation}
The following computing algorithm can be employed to solve (\ref{eqn:cond_pls0}).


\begin{center}
\begin{minipage}[c]{0.8\textwidth}
\hrule
\medskip
{\it Algorithm 1}: \\
{\it  Step 1.} Initialize $\widehat \bB^{(0)}$, $u_{jk}$ and $v_{st}$. \\
{ \it Step 2.} For $k=1,\ldots,q$, solve (\ref{eqn:cond_pls}) for $\hat \beta_k$ and $\hat \gamma_k$. 
\medskip 
\hrule
\end{minipage}
\end{center}

As computational remarks, $\widehat \bB^{(0)}$ can be initialized by the separate Lasso regression ignoring the dependency structure. The weights $u_{jk}$ and $v_{sk}$ are set as $|\tilde \beta_{jk}|^{-1}$ and $|\tilde \gamma_{sk}|^{-1}$ as in Zou (2006), where $\tilde \beta_{jk}$ and $\tilde \gamma_{sk}$ are any consistent estimates of $\beta_{jk}$ and $\gamma_{sk}$, respectively. Since (\ref{eqn:cond_pls}) is a convex optimization problem, its global minimum can be obtained by any available adaptive Lasso regression procedure. Furthermore, the coordinate descent algorithm (Friedman et al., 2007) can be employed to further improve the computational efficiency of solving (6). More importantly, {\it Step 2} fits the adaptive Lasso regression model (6) for each $k$, and thus can be easily parallelized and distributed to multiple computing nodes. Therefore, {\it Algorithm 1} is scalable and can efficiently handle dataset with big size.


When identifying the sparsity in the conditional graphical model defined by $\bOmega$, the symmetry of $\bOmega$ implies that $\sign(\omega_{sk})=\sign(\omega_{ks})$, and thus $\sign(\gamma_{sk})=\sign(\gamma_{ks})$. Consequently, additional refinement is necessary to correct the possible inconsistency in $\sign(\widehat \gamma_{sk})$. Similar as in Meinshausen and B$\ddot{\rm u}$hlmann (2006), one natural way is to set
$$
\widehat \gamma^{\wedge}_{sk}=0~\mbox{if}~\widehat \gamma_{sk}=0 \wedge \widehat \gamma_{ks}=0,
$$
or a less conservative way is to set
$$
\widehat \gamma^{\vee}_{sk}=0~\mbox{if}~\widehat \gamma_{sk}=0 \vee \widehat \gamma_{ks}=0.
$$
In the numerical experiments, the less conservative way is used and the resultant selection performance in $\widehat{\Omega}$ appears to be satisfactory.

\section{Asymptotic properties}


This section establishes the asymptotic properties of the proposed multivariate conditional regression model in terms of the selection and estimation accuracy. Let ${\bB}^*=(\beta^*_{jk})$ be the true regression coefficient matrix, $\bOmega^*=(\omega^*_{sk})$ be the inverse of the true covariance matrix $\bSigma^*=(\sigma^*_{sk})$, and $\bGamma^*=(\gamma^*_{sk})$ be defined as in (\ref{eqn:spar_Omega}) with $\bOmega^*$. The selection accuracy is measured by the sign agreement between $(\widehat{\bB},\widehat{\bOmega})$ and $({\bB}^*,{\bOmega}^*)$, and the estimation accuracy is quantified by the asymptotic normality of $n^{1/2}(\widehat{\bB}-{\bB}^*)$.

Without loss of generality, we assume that $\sigma^*_{ss}=1$ for all $s$'s, and denote 
$$
{\bM} = \left( \begin{array}{cc} n^{-1} {\bX}^T \bX & 0 \\ 0 & \bSigma^* \end{array} \right).
$$
Let ${\cal A}^{\beta}_k = \{j: \beta^*_{jk} \neq 0 \}$, ${\cal A}^{\beta} = \{(j,k): j \in {\cal A}^{\beta}_k \}$, ${\cal A}^{\omega}_k = \{s: s \ne k, \omega^*_{sk} \neq 0\}=\{s: \gamma^*_{sk} \neq 0\}={\cal A}^{\gamma}_k$, ${\cal A}^{\omega} = \{(s,k): s \in {\cal A}^{\omega}_k \}$, ${\cal A}={\cal A}^{\beta} \cup {\cal A}^{\omega}$, and ${\cal A}_k = \{j: j \in {\cal A}^{\beta}_k\} \cup \{p+s: s \in {\cal A}^{\omega}_k\}$. Let $d^{\beta}_k=|{\cal A}^{\beta}_k|$,  $d^{\omega}_k=|{\cal A}^{\omega}_k|$, $d_k=|{\cal A}_k|$, and $d=\max_k\{d_k\}$. Denote $\Lambda_{min}(A)$ and $\Lambda_{max}(A)$ as the minimum and maximum eigenvalues of a matrix $A$. Denote $\lambda_{init}$ and $\lambda_{1n}=\lambda_{2n}=\lambda_n$ as the tuning parameters used in the initial Lasso regression and (\ref{eqn:cond_pls}), respectively. The following technical conditions are assumed.
\begin{enumerate}
\item[(A1)] There exists a positive constant $a_1$ such that $\Lambda_{max}(n^{-1}{\bX}^T {\bX}) \leq a_1$ and $\Lambda_{max}(\bSigma^*) \leq a_1$. In addition, $n^{-1/2} \max_i \{ {\bx}_i^T {\bx}_i\} \rightarrow 0$.
\item[(A2)] For some integers $1 \leq d \leq (p+q)/2$, $m \geq d$, $m+d \leq p+q$ and a positive constant $k_0$,
$$
\frac{1}{K(d,m,k_0,\bM)} := 
\min_{J_0 \subset \{1,\ldots,p+q\}, |J_0| \leq d} \left (  \min_{\alpha \neq 0, \|\alpha_{J_0^c}\|_1 \leq k_0 \|\alpha_{J_0}\|_1 } \left ( \frac{ \| {\bM}^{1/2} \alpha \|_2}{\|\alpha_{J_{0m}}\|_2} \right ) \right )>0.
$$
\item[(A3)] Let ${\zeta}^*_{min} = \min_{(j,k) \in {\cal A}}( |\beta^*_{jk}|,|\omega^*_{jk}|)$ and $\Lambda_{min}(d)$ be defined as below, 
\begin{equation}
\begin{split}
(n{\zeta}^*_{min}\Lambda_{min}(d))^{-1} & O \big(\max \big( d \lambda_{init} (\Lambda_{min}(d))^{1/2} K(d,d,3,\bM)^2, \lambda_n d^{1/2} (\Lambda_{min}(d))^{-1}, \\
& n^{-1} d^{1/2} \lambda_{init}, n^{1/2} d^{1/2} (\log (p+q))^{1/2},n^{-1} d \lambda_{init}^{-2} K(d,d,3,\bM) \big ) \big) \rightarrow 0.
\nonumber
\end{split}
\end{equation}

\end{enumerate}

Assumption (A1) implies that $\Lambda_{max}(\bM) \leq a_1$, $\max_j \{n^{-1} {\bx}^j ({\bx}^j)^T\} \leq a_1$, and $\min_k \{\omega^*_{kk}\} \geq \Lambda_{min}(\Omega^*) \geq 1/a_1$. Assumption (A2) is similar as the restricted eigenvalue assumption in Bickel et al. (2008) and Zhou et al. (2009). It implies that for any subset $S \subset \{1,\ldots,p+q\}$ with $|S| \leq d$, we have $\Lambda_{min}({\bM}_{SS}) \geq \Lambda_{min}(d)>0$, where
$$
\Lambda_{min}(d)=\min_{J_0 \subset \{1,\ldots,p+q\}, |J_0| \leq d} \left (  \min_{\alpha \neq 0, \alpha_{J_0^c}=0 } \left ( \frac{ \| \alpha^T {\bM} \alpha \|_2}{\alpha_{J_0}^T \alpha_{J_0}} \right ) \right ).
$$
Assumption (A3) is similar as the condition in Zhao and Yu (2006) and Meinshausen (2007), and implies that the nonzero $\beta^*_{jk}$ and $\gamma^*_{sk}=-\omega^*_{sk}/\omega^*_{kk}$ will not decay too fast to be dominated by the noise terms. 


\begin{theorem}
\label{thm:sel_cons}
(Selection consistency) Supposed that conditions (A1)-(A3) are satisfied with $m=d$ and $k_0=3$, the initial $\tilde \beta_{jk}$ and $\tilde \gamma_{sk}$ are set as the solution of the separate Lasso regression, and $\lambda_{1n}=\lambda_{2n}=\lambda_n$.  Then as $n \rightarrow \infty$,
$$
P(\sgn(\widehat{\bB}) \neq \sgn({\bB}^*)~\mbox{or}~\sgn(\widehat{\bOmega}) \neq \sgn({\bOmega}^*))\longrightarrow 0,
$$
when $n^{-1/2}d \lambda_{init} \rightarrow 0$,  $\min\big( n^{-3} \lambda_n^2 d \lambda_{init}^2 K(d,d,3,\bM)^4,n \Lambda_{min}(d) (\zeta^*_{min})^2 \big ) (\log(p+q))^{-1}  \rightarrow \infty$, and $(n\Lambda_{min}(d))^{-1} \max( \lambda_{init} d,n^{-1/2}\lambda_{init}d(\log (p+q))^{1/2}, n^{-1} \lambda_{init}^2d) \rightarrow 0$.
\end{theorem}
 
\begin{theorem}
\label{thm:est_cons}
(Asymptotic normality) Supposed that the conditions in Theorem \ref{thm:sel_cons} are satisfied. Let $s_k^2=\tilde \sigma_{kk}^* {\alpha}^T {\bM}_{{\cal A}_k,{\cal A}_k}^{-1} \alpha$, where $\alpha$ is any $|{\cal A}_k| \times 1$ vector with unit length, and ${\bM}_{{\cal A}_k,{\cal A}_k}$ is the principle submatrix of $\bM$ defined by ${\cal A}_k$. Then 
$$
n^{1/2} s_k^{-1} {\alpha}^T \left( \left ( \begin{matrix} \widehat{\beta}_k \\ \widehat{\gamma}_k \end{matrix} \right ) - \left ( \begin{matrix} {\beta}^*_k \\ {\gamma}^*_k \end{matrix} \right ) \right ) \stackrel{d}{\longrightarrow} N(0,1) ~~\mbox{for any}~k,
$$
when $d \lambda_n (p+q)^{-1} \rightarrow 0$, $n^{-1/2} \lambda_n d^{1/2} (\Lambda_{min}(d) \zeta^*_{min})^{-1} \rightarrow 0$ and $n^{-1/2} \lambda_{init} d \Lambda_{min}(d))^{-1/2} \rightarrow 0$. 
\end{theorem}

Theorems 1 and 2 show that with consistent initial estimates of $\bB$ and $\bOmega$, the proposed multivariate conditional regression model is able to achieve both selection consistency and the asymptotic normality. 


\section{Numerical experiments}

This section examines the effectiveness of the proposed multivariate conditional regression model on a variety of simulated examples and a real application to the Glioblastoma Cancer Dataset (TCGA, 2008). The proposed multivariate conditional regression model with the adaptive Lasso penalty, denoted as aMCR, is compared against 
the alternative updating algorithm in (\ref{eqn:multi_pls}) (ALT; Yin and Li, 2011; Lee and Liu, 2012), and the separate Lasso regression (SEP; estimating each $\beta_k$ and $\Omega$ separately). 

The comparison is conducted with respect to the estimation and selection accuracy of $\widehat{\bB}$ and $\widehat{\bOmega}$. In specific, the estimation accuracy of $\widehat{\bB}$ is measured by the Frobenius norm $\|\bDelta_B\|_F=\big(\sum_{i,j} (\bDelta_B)_{ij}^2 \big)^{1/2}$, the matrix 1-norm $\|\bDelta_B\|_1=\max_j \sum_i |(\bDelta_B)_{ij}|$, and the matrix $\infty$-norm $\|\bDelta_B\|_{\infty}=\max_i \sum_j |(\bDelta_B)_{ij}|$, where $\bDelta_{B}=\widehat{\bB}-{\bB}^*$. The estimating accuracy of $\widehat{\bOmega}$ is not reported as the primary interest is the sparsity inferred by $\widehat{\bOmega}$, and the proposed method does not produce $\widehat{\bOmega}$ directly. The selection accuracy of $\widehat{\bB}$ and $\widehat{\bOmega}$ is measured by the symmetric difference 
\begin{eqnarray*}
\Dist(\widehat{\cA}^{\beta}, {\cA}^{\beta}) &=& \frac{\big|\widehat{\cA}^{\beta} \char`\\ {\cA}^{\beta} \big|+ \big|{\cA}^{\beta} \char`\\ \widehat{\cA}^{\beta} \big|}{pq}; \\ \Dist(\widehat{\cA}^{\omega}, {\cA}^{\omega}) &=& \frac{\big|\widehat{\cA}^{\omega} \char`\\ {\cA}^{\omega} \big|+ \big|{\cA}^{\omega} \char`\\ \widehat{\cA}^{\omega} \big|}{q^2},
\end{eqnarray*}
where $\widehat{\cA}^{\beta}$ and $\widehat{\cA}^{\omega}$ are the active sets defined by $\widehat{\bB}$ and $\widehat{\bOmega}$, and $|\cdot|$ denotes the set cardinality. We also report the specificity (Spe), sensitivity (Sen) and Matthews correlation coefficient (Mcc) scores, defined as
\begin{eqnarray*}
\Spe &=& \frac{\TN}{\TN+\FP},~~~\Sen = \frac{\TP}{\TP+\FN}, \\
\Mcc &=& \frac{\TP \times \TN - \FP \times FN}{\sqrt{(\TP+\FN)(\TN+\FP)(\TP+\FP)(\TN+\FN)}},
\end{eqnarray*}
where TP, TN, FP and FN are the numbers of true positives, true negatives,
false positives and false negatives in identifying the nonzero elements in $\widehat{\bB}$ or $\widehat{\bOmega}$, and ``positive" refers to the nonzero entries. 

Furthermore, tuning parameters are used in most penalized log-likelihood formulations to balance the model estimation and model complexity. For example, the tuning parameters $\lambda_{1n}$ and $\lambda_{2n}$ in (\ref{eqn:multi_pls}) and  (\ref{eqn:cond_pls}) control the tradeoff between the sparsity and the estimation accuracy of the multivariate regression models. In the numerical experiments, we employed Bayesian information criterion (BIC; Schwarz, 1978) to select the tuning parameters, which is shown to perform well in tuning penalized likelihood method (Want et al., 2007). The BIC criterion is minimized through a grid search on a two-dimensional equally-spaced grid $(10^{-3+(s-1)/3},10^{-3+(t-1)/3})$; $s,t=1,\ldots,19$. Other data adaptive model selection criteria, such as cross validation, can be employed as well (Lee and Liu, 2012).

\subsection{Simulated examples}

The simulated examples follow the same setup as in Li and Gui (2006), Fan et al. (2009), Peng et al. (2009) and Yin and Li (2011). First, each entry of the precision matrix $\bOmega$ is generated from the product of a Bernoulli random variable with success rate proportional to $1/q$ and a uniform random variable on $[-1,-0.5]\cup[0.5, 1]$. For each row, all off-diagonal entries are divided by the sum of the absolute value of the off-diagonal entries multiplied by 3/2. The final precision matrix $\bOmega$ is obtained by symmetrizing the generated matrix and setting the diagonal entries as 1. Next, each entry of the coefficient matrix $\bB$ is generated from the product of a Bernoulli random variable with success rate proportional to $1/p$ and a uniform random variable on $[-1,-v_m]\cup[v_m, 1]$, where $v_m$ is the minimum absolute value of the nonzero entries in $\bOmega$. Finally, with the generated $\bOmega$ and $\bB$, each entry of the covariate matrix $\bX$ is generated independently from $\mbox{Bern}(1/2)$, and the response vector is generated from $Y|X=x \sim N_q({\bB}^T x, {\bOmega}^{-1})$. 

Six models are considered, and for each given model, a training sample of $n$ observations $({\bx}_i,{\by}_i);~i=1,\ldots,n$ are generated. 

Model 1: $(p, q,n) = (100, 100, 250)$, where $P({\bB}_{ij} \neq 0) = 3/p$ and $P( {\bOmega}_{ij} \neq 0) = 2/q$;

Model 2: $(p, q,n) = (50, 50, 250)$, where $P({\bB}_{ij} \neq 0) = 4/p$ and $P({\bOmega}_{ij} \neq 0) = 2/q$;

Model 3: $(p, q,n) = (10, 25, 250)$, where $P({\bB}_{ij} \neq 0) = 3.5/p$ and $P({\bOmega}_{ij} \neq 0) = 2/q$;

Model 4: $(p, q,n)=(200, 1000, 250)$, where $P({\bB}_{ij} \neq 0) = 20/p$ and $P({\bOmega}_{ij} \neq 0) = 1.5/q$;

Model 5: $(p, q,n)=(200, 800, 250)$, where $P({\bB}_{ij} \neq 0) = 25/p$ and $P({\bOmega}_{ij} \neq 0) = 1.5/q$;

Model 6: $(p, q,n)=(200, 400, 150)$, where $P({\bB}_{ij} \neq 0) = 20/p$ and $P({\bOmega}_{ij} \neq 0) = 2.5/q$.

Each model is replicated 50 times, and the averaged performance measures as well as the estimated standard errors are reported in Tables \ref{tab:simB} and \ref{tab:simO}.
 
\vspace{-0.5cm}
\begin{center}
\begin{tabular}[t]{c}
\hline \hline
Tables \ref{tab:simB} and \ref{tab:simO} about here \\
\hline \hline
\end{tabular}
\end{center}

It is evident that the proposed aMCR delivers superior numerical performance, in terms of both estimation and selection accuracy of $\bB$ and $\bOmega$, against other competitors across all six simulated examples. In Tables \ref{tab:simB} and \ref{tab:simO}, we only report the numerical performance of ALT on examples 2 and 3, due to the computational burden of running ALT on other examples with larger dimensions. Although the performance of ALT might be improved if some random start algorithm is employed to partially overcome the issue of local minimum, the inefficient alternating algorithm becomes one major obstacle of applying ALT to analyze high dimensional dataset.

In Tables \ref{tab:simB} and \ref{tab:simO}, the advantage of aMCR and ALT over SEP demonstrates that inclusion of the covariance matrix in (\ref{eqn:multi_pls}) and (\ref{eqn:cond_pls}) is indeed helpful in identifying the sparsity in $\bB$ and $\bOmega$ and thus in estimating $\bB$. As for the selection accuracy, aMCR yields higher Spe and Mcc but lower Sen in most examples. This is due to the fact that aMCR tends to produce sparser models than SEP since the correlations among the responses are positive (Lee and Liu, 2012). Although sparser models are produced, aMCR still yields smaller symmetric difference than SEP. As for the estimation accuracy of $\bB$, it is clear that aMCR outperforms SEP under all three metrics of $\widehat{\bB}-{\bB}^*$. This implies that the proposed multivariate conditional regression model can improve not only the accuracy of identified nonzero entries in the precision matrix, but also the accuracy of estimating the multivariate regression coefficient matrix.

\subsection{Real application}

In this section, we apply the proposed multivariate conditional regression model to a Glioblastoma multiforme (GBM) cancer dataset studied by the Cancer Genome Atlas (TCGA) Research Network (TCGA, 2008; Verhaak et al., 2010). GBM is the most common and most aggressive malignant primary brain tumor in adults. The original dataset collected by TCGA 
consists of 202 samples, 11861 gene expression values and 534 microRNA expression values. One primary goal of the study is to regress the microRNA expressions on the gene expressions and model how the microRNAs regulate the gene expressions. It is also of interest to construct the underlying network among the microRNAs. The proposed model can achieve these two goals simultaneously, where the sparse coefficient matrix reveals the regulatory relationship among the microRNA and gene expressions and the sparse precision matrix can be interpreted as the dependency structure among the microRNAs. 

For illustration, some preliminary data cleaning is conducted by removing missing values and prescreening the less expressed genes and microRNAs as in TCGA et al. (2010) and Lee and Liu (2012). In particular, 6 samples with missing values are removed, and thus 196 complete samples are remained in the dataset. Furthermore, the genes and microRNAs are sorted based on their corresponding median absolute deviation (MAD), and the top 500 genes and top 20 microRNAs with large MADs are selected. 

The dataset is randomly split into a training set with 120 samples and a test set with 76 samples. On the training set, each method is fitted to estimate the multivariate regression coefficient matrix and the precision matrix.  Since the truth is unknown in the real application, the estimation performance is measured by the predictive square error (Pse) estimated on the test set, defined as
$$
\mbox{Pse}=|\mbox{test set}|^{-1} \sum_{\mbox{test set}} \|{\bY}_i - \widehat{\bY}_i \|_F^2,
$$
where $|\mbox{test set}|$ denotes the cardinality of the test set. In addition, the numbers of the selected genes by each method are also reported.

The averaged Pse and numbers of selected genes as well as their estimated standard errors based on 50 replications are reported in Table \ref{tab:gene}.

\vspace{-0.5cm}
\begin{center}
\begin{tabular}[t]{c}
\hline \hline
Table \ref{tab:gene} about here \\
\hline \hline
\end{tabular}
\end{center}

Clearly, the proposed aMCR yields sparser multivariate regression model and achieves smaller Pse than the separate regression model. This agrees with the conclusion in Lee and Liu (2012), and the sparser regression model is due to the fact that the joint estimation method is able to obtain more shrinkage when strong positive correlations are present among the selected microRNAs.  Again, the numerical performance of ALT is not reported due to the computational burden, but we note that in Lee and Liu (2012), the Pse of the ALT is 1.23(.032) and the number of selected genes is 78.0(32.15) based on a slightly smaller dataset.

\begin{figure}[h]
\caption{The dependency network of the selected microRNAs based on the estimated sparse precision matrix.}
\begin{center}
\includegraphics[width=4in,height=4in]{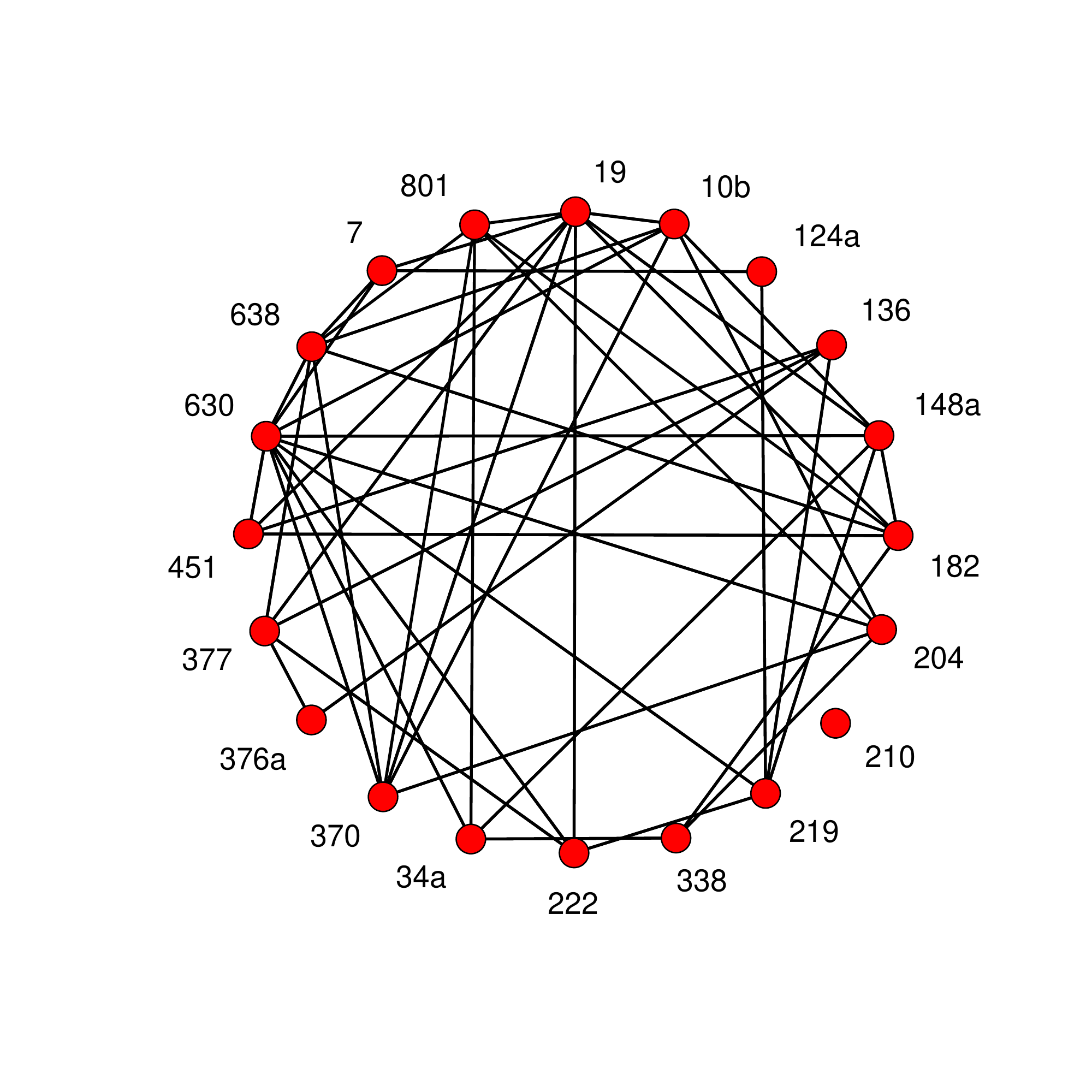}
\end{center}
\label{fig:gene}
\end{figure}

Figure \ref{fig:gene} displays the estimated conditional dependency structure among the microRNAs based on the estimated precision matrix of the microRNAs. Compared with the results in Lee and Liu (2012), the graphical structure in Figure \ref{fig:gene} captures the strong positive correlations among the selected microRNA pairs, including the tuple of hsa.mir.136, hsa.mir.376a and hsa.mir.377. More importantly, it produces a sparser dependency structure than that in Lee and Liu (2012), and rules out more microRNA pairs with weak correlations, such as hsa.mir.bart19 and hsa.mir.124a (with pairwise correlation $-0.12$).

\vspace{-0.5cm}
\begin{center}
\begin{tabular}[t]{c}
\hline \hline
Figure \ref{fig:gene} about here \\
\hline \hline
\end{tabular}
\end{center}

\section{Summary}

This article proposes a joint estimation method for estimating the multivariate regression model and the dependency structure among the multiple responses. As opposed to the existing methods maximizing the penalized joint log-likelihood function, the proposed method is formulated as a penalized conditional log-likelihood function, leading to efficient computation and superior numerical performance. Its asymptotic estimation and selection consistencies are established for diverging dimensions and numbers of responses. 
Finally, it is worth pointing out that the penalized conditional log-likelihood formulation can be extended to a general framework without the Gaussian distributional assumption such as in Finegold and Drton (2011) and Lee et al. (2012).

\section*{Acknowledgment}
The author would like to thank Wonyul Lee and Yufeng Liu (University of North Carolina at Chapel Hill) for sharing their code on the alternative updating algorithm and the Glioblastoma multiforme cancer dataset.

\section*{Appendix}

\noindent{\bf Proof of Theorem \ref{thm:sel_cons}:} We first establish upper bounds for $P(\sgn(\hat \beta_k) \neq \beta^*_k)$ and $P(\hat \gamma_k \neq \gamma^*_k)$, where $\hat \beta_k$ and $\hat \gamma_k$ are the solution of 
\begin{equation}
\min_{\beta_k,\gamma_k}~\|{\by}^{k} - {\bX} \beta_k - \widehat {\by}^{-k} \gamma_k\|^2 + \lambda_n \Big ( \sum_{j=1}^p u_{jk} |\beta_{jk}| + \sum_{s \neq k} v_{sk} |\gamma_{sk}| \Big ),
\label{eqn:cond_pls2}
\end{equation}
and $\widehat {\by}^{-k}={\by}^{-k}-\bX \widehat{\bB}_{-k}^{(0)}$ is a surrogate of ${\be}^{-k}={\by}^{-k}-\bX {\bB}^*_{-k}$. Based on the model assumption (\ref{eqn:cond_dist}), we have
\begin{equation}
{\by}^k={\bX} \beta^*_k + {\widehat \by}^{-k} \gamma^*_k  + {\bxi}_k + {\bepsilon}_k,
\label{eqn:cond_dist2}
\end{equation}
where ${\bxi}_k = ({\be}^{-k} - {\widehat \by}^{-k}) \gamma^*_k=\bX(\widehat{\bB}_{-k}^{(0)}-{\bB}^*_{-k})\gamma^*_k$ and ${\bepsilon}_k \sim N({\bf 0}_n,\tilde{\sigma}^*_{kk} \bI_n)$. Furthermore, let $\zeta=(\beta_k^T, \gamma_k^T)^T$ be the augmented coefficient vector, $\bZ=(\bX,{\widehat \by}^{-k})$ be the augmented covariate matrix, $r=(u_k^T,v_k^T)^T$, and then the model (\ref{eqn:cond_pls2}) can be simplified as
\begin{equation}
\min_{\widetilde{\beta}}~\|{\by}^{k} - {\bZ} \zeta \|^2 + \lambda_n  \sum_{j=1}^{p+q-1} r_j |\zeta_j|.
\label{eqn:cond_pls3}
\end{equation}

We now verify the conditions (\ref{eqn:eigenbound}) and (\ref{eqn:irrep}) in Lemma \ref{lem:signrec}. First, for simplicity, let
$$
{\cal T}=\Big \{ \max_{j,s}~n^{-1} ({\bX}^j)^T {\be}^s \leq a_1^2 (8n^{-1} \log(p+q))^{1/2} \Big \},
$$
and it follows from the proof of Lemma 9.1 in Zhou et al. (2009) that $P({\cal T}) \geq 1-(p+q)^{-2}$. Also let $\widetilde{\bZ}=(\bX,{\be}^{-k})$, ${\bM}^{-k}$ be the submatrix of $\bM$ without the $(p+k)$-th row and column, $\bDelta=n^{-1} \widetilde{\bZ}^T \widetilde{\bZ} - {\bM}^{-k}$, and
$$
{\cal Y}_k=\Big \{ \max_{j,s} |{\bDelta}_{js}| \leq 8 n^{-1/2} (\log (p+q))^{1/2} \Big \}.
$$
It then follows from Lemma 9.3 in Zhou et al. (2009) that $P({\cal Y}_k) \geq 1-(p+q)^{-2}$. Additionally, since $\Lambda_{min}(d)$ is asymptotically larger than $n^{-1/2} (\log (p+q))^{1/2}$, there exists a constant $c_1>0$ such that on the set ${\cal T} \cap {\cal Y}_k$,
$$
\Lambda_{min} \big( n^{-1} \widetilde{\bZ}_{{\cA}_k}^T \widetilde{\bZ}_{{\cA}_k} \big) \geq 2 c_1 \Lambda_{min}(d),
$$
for any subset ${\cA} \subset \{1,\ldots,p+q\}{\char`\\}\{p+k\}$ with $|{\cA}| \leq d$.
Furthermore, let ${\cA}_{1}=\{1 \leq j \leq p: j \in {\cA}\}$ and ${\cA}_{2}=\{1 \leq s \leq q: p+s \in {\cA}\}$, then 
\begin{eqnarray*}
& & \big | \Lambda_{min}( n^{-1} {\bZ}_{{\cA}}^T {\bZ}_{{\cA}}) - \Lambda_{min}( n^{-1} \widetilde{\bZ}_{{\cA}}^T \widetilde{\bZ}_{{\cA}}) \big | \\
&\leq& \|n^{-1} {\bZ}_{{\cA}}^T {\bZ}_{{\cA}}-n^{-1} \widetilde{\bZ}_{{\cA}}^T \widetilde{\bZ}_{{\cA}} \|_2 \leq \|n^{-1} {\bZ}_{{\cA}}^T {\bZ}_{{\cA}}-n^{-1} \widetilde{\bZ}_{{\cA}}^T \widetilde{\bZ}_{{\cA}} \|_{\infty} \\
&\leq& \|n^{-1} \widehat{\by}_{{{\cA}_{2}}}^T {\bX}_{{{\cA}_{1}}} \|_{\infty} + \|n^{-1} {\bX}_{{{\cA}_{1}}}^T \widehat{\by}_{{{\cA}_{2}}} \|_{\infty} + \|n^{-1} \widehat{\by}_{{{\cA}_{2}}}^T \widehat{\by}_{{{\cA}_{2}}}-n^{-1} {\be}_{{{\cA}_{2}}}^T {\be}_{{{\cA}_{2}}}\|_{\infty},
\end{eqnarray*}
where $\|M\|_2$ is the operator norm of a matrix $M$, and $\|M\|_{\infty}=\max_i \sum_j |M_{ij}|$. But since $\widehat {\by}^{-k}={\by}^{-k}-\bX \widehat{\bB}_{-k}^{(0)}$ with $\widehat{\bB}_{-k}^{(0)}$ being the Lasso estimate, we have on the set ${\cal T}$,
$$
\max(\|n^{-1} \widehat{\by}_{{{\cA}_{2}}}^T {\bX}_{{{\cA}_{1}}} \|_{\infty},\|n^{-1} {\bX}_{{{\cA}_{1}}}^T \widehat{\by}_{{{\cA}_{2}}} \|_{\infty}) \leq O(n^{-1} \lambda_{init} d).
$$ 
Also, conditional on the set ${\cal T}$, it follows from Assumption (A1) that
$$
\|n^{-1} \widehat{\by}_{{{\cA}_{2}}}^T \widehat{\by}_{{{\cA}_{2}}}-n^{-1} {\be}_{{{\cA}_{2}}}^T {\be}_{{{\cA}_{2}}}\|_{\infty} \leq O(n^{-1} d\lambda_{init} n^{-1/2}(\log (p+q))^{1/2}) + O(n^{-2} d \lambda_{init}^2 ).
$$ 
Therefore, on the set ${\cal T} \cap {\cal Y}_k$, 
\begin{eqnarray*}
\Lambda_{min} \big( n^{-1} {\bZ}_{{\cA}_k}^T {\bZ}_{{\cA}_k} \big) &\geq& \Lambda_{min} \big( n^{-1} \widetilde{\bZ}_{{\cA}_k}^T \widetilde{\bZ}_{{\cA}_k} \big)-\big | \Lambda_{min}( n^{-1} {\bZ}_{{\cA}}^T {\bZ}_{{\cA}}) - \Lambda_{min}( n^{-1} \widetilde{\bZ}_{{\cA}}^T \widetilde{\bZ}_{{\cA}}) \big | \\
&\geq& c_1 \Lambda_{min}(d),
\end{eqnarray*}
for sufficiently large $n$, since the cardinality of ${\cA}_k$ is bounded by $d$ and $\Lambda_{min}(d)$ is asymptotically larger than $\max(n^{-1} \lambda_{init} d,n^{-3/2}\lambda_{init}d(\log (p+q))^{1/2}, n^{-2} \lambda_{init}^2d)$.

Next, it follows from Bickel et al. (2008) that under Assumption (A2) with $m=d$ and $k_0=3$,
\begin{eqnarray*}
\delta_{{\cA}_k} := \max_{j \in {\cA}_k}~| \tilde \zeta_j - \zeta^*_j | &\leq& 4K(d,d,3,\bM)^2 n^{-1} d^{1/2} \lambda_{init}; \\
\delta_{{\cA}_k^c} := \max_{j \in {\cA}_k^c}~| \tilde \zeta_j - \zeta^*_j | &\leq& 16K(d,d,3,\bM)^2 n^{-1} d^{1/2} \lambda_{init},
\end{eqnarray*}
on set $\cal T$, where $\tilde \zeta_j$ is the solution of the Lasso regression. Therefore,
$$
\frac{r_{min}({{\cA}_k^c})}{r_{max}({{\cA}_k})} = \frac{\min_{j \in {\cA}_k} |\tilde \zeta_j|}{\max_{j \in {\cA}_k^c} |\tilde \zeta_j|} \geq \frac{ \zeta^*_{min} - \delta_{{\cA}_k}}{\delta_{{\cA}_k^c}},
$$
where ${\zeta}^*_{min} = \min_{j \in {\cA}_k} |\zeta^*_j|$. Furthermore, it follows from Lemma 10.3 of Zhou et al. (2009) that on set ${\cal Y}_k$, there exists a positive constant $c_2$ such that $\| {\bZ}_{{\cA}_k^c}^T {\bZ}_{{\cA}_k}({\bZ}_{{\cA}_k}^T {\bZ}_{{\cA}_k})^{-1} \|_{\infty} \leq  c_2 d^{1/2} (\Lambda_{min}(d))^{-1/2}$. Therefore, on the set ${\cal T} \cap {\cal Y}_k$, when $n$ is sufficiently large,
$$
\| {\bZ}_{{\cA}_k^c}^T {\bZ}_{{\cA}_k}({\bZ}_{{\cA}_k}^T {\bZ}_{{\cA}_k})^{-1} \|_{\infty} \leq \frac{r_{min}({{\cA}_k^c})}{r_{max}({{\cA}_k})}(1-\eta),
$$
for some $0 < \eta <1$, provided that $(\zeta^*_{min})^{-1} n^{-1} d \lambda_{init} (\Lambda_{min}(d))^{-1/2} K(d,d,3,\bM)^2 \rightarrow 0$.

Finally,it follows from Lemma \ref{lem:signrec} that for each $k=1,\ldots,q$,
$$
P(\sgn(\widehat{\zeta}) \neq \sgn({\zeta}^*)) = O((p+q)^{-2}),
$$
provided that $n^{-1}d \lambda_{init} \rightarrow 0$ and $\min\big( n^{-1} \lambda_n^2 r_{min}^2({\cA}_k^c),n \Lambda_{min}(d) (\zeta^*_{min})^2 \big ) (\log(p+q))^{-1}  \rightarrow \infty$. Consequently, $P(\sgn(\widehat{\bB}) \neq \sgn(\bB)~\mbox{or}~\sgn(\widehat{\bOmega}) \neq \sgn(\bOmega)) \leq q O((p+q)^{-2})$, which implies the desired result immediately.

\begin{lemma}
\label{lem:signrec}
Consider the linear model in (\ref{eqn:cond_pls3}), where the design matrix $\bZ$ satisfies 
\begin{eqnarray}
\Lambda_{min}\big( n^{-1} {\bZ}_{{\cA}_k}^T {\bZ}_{{\cA}_k} \big) &\geq& c_1 \Lambda_{min}(d)>0, \label{eqn:eigenbound}\\
\| {\bZ}_{{\cA}_k^c}^T {\bZ}_{{\cA}_k}({\bZ}_{{\cA}_k}^T {\bZ}_{{\cA}_k})^{-1} \|_{\infty} &\leq& \frac{r_{min}({{\cA}_k^c})}{r_{max}({{\cA}_k})}(1-\eta) \label{eqn:irrep}
\end{eqnarray}
for $c_1>0$ and $0<\eta<1$, where $r_{min}({{\cA}_k^c})=\min_{j \in {{\cA}_k^c}} r_j$, and $r_{max}({{\cA}_k})=\max_{j \in {{\cA}_k}} r_j$. Let ${\zeta}^*_{min} = \min_j |\zeta^*_j|$ be asymptotically larger than $
(\Lambda_{min}(d))^{-1} O \big(\max \big(n^{-1} \lambda_n d^{1/2} r_{max}({{\cA}_k}),$ $n^{-1} d^{1/2} \lambda_{init}, n^{-1/2} d^{1/2} (\log (p+q))^{1/2}, n^{-2} d \lambda_{init}^{-2} K(d,d,3,\bM) \big ) \big)$.
Then
$$
P(\sgn(\widehat{\zeta}) \neq \sgn({\zeta}^*)) = O((p+q)^{-2}),
$$ 
provided that $n^{-1/2}d \lambda_{init} \rightarrow 0$ and $\min\big( n^{-1} \lambda_n^2 r_{min}^2({\cA}_k^c),n \Lambda_{min}(d) (\zeta^*_{min})^2 \big ) (\log(p+q))^{-1}  \rightarrow \infty$.
\end{lemma}

\noindent{\bf Proof of Lemma \ref{lem:signrec}:} Denote ${\bz}^j$ as the $j$-th column of $\bZ$. It follows from the Karush-Kuhn-Tucker condition that $\widehat{\zeta}$ must satisfy that 
\begin{eqnarray}
({\bz}^j)^T ({\by}^{k} - {\bZ} \zeta) = \lambda_n r_j \sign(\zeta_j), &&\mbox{if}~\widehat{\zeta}_j \neq 0; \label{eqn:kkt1} \\
\big | ({\bz}^j)^T ({\by}^{k} - {\bZ} \zeta)\big | \leq \lambda_n r_j, &&\mbox{if}~\zeta_j = 0. \label{eqn:kkt2}
\end{eqnarray}
Consider the following equation based on ${\bZ}_{{\cA}_k}$, 
$$
{\bZ}_{{\cA}_k}^T {\by}^{k} - {\bZ}_{{\cA}_k}^T {\bZ}_{{\cA}_k} \bar{\zeta}_{{\cA}_k} = \lambda_n \bar{\bs}_{{\cA}_k}
$$
where $\bar{\bs}_{{\cA}_k}=\big(r_j \sign(\zeta^*_j);j \in {{\cA}_k} \big )$. By (\ref{eqn:cond_dist2}), the solution to the above equation is
\begin{equation}
\bar{\zeta}_{{\cA}_k} = \zeta^*_{{\cA}_k} +  ({\bZ}_{{\cA}_k}^T {\bZ}_{{\cA}_k})^{-1}  ({\bZ}_{{\cA}_k}^T ({\bxi}_k+{\bepsilon}_k) - \lambda_n {\bar{\bs}}_{{\cA}_k}).
\label{eqn:cond_sol1}
\end{equation}

Note that if $\sgn(\bar \zeta_{{\cA}_k}) = \sgn(\zeta^*_{{\cA}_k})$, the following $\widehat \zeta$ with $(\widehat \zeta_j)_{j \in {{\cA}_k}} = \bar \zeta_{{\cA}_k},~(\widehat \zeta_j)_{j \notin {{\cA}_k}} = 0$ is a solution of (\ref{eqn:kkt1})-(\ref{eqn:kkt2}). Therefore, $\sgn(\widehat \zeta)=\sgn(\zeta^*)$ if
$$
\sgn(\bar \zeta_{{\cA}_k}) = \sgn(\zeta^*_{{\cA}_k}),~\mbox{and}~\big | ({\bz}^j)^T ({\by}^{k} - {\bZ}_{{\cA}_k} \bar{\zeta}_{{\cA}_k})\big | \leq \lambda_n r_j, ~\mbox{if}~j \notin {\cal A}_k.
$$
This statement is similar to Proposition 1 of Zhao and Yu (2006) and (S.5) of Huang et al. (2008). It implies that
$$
P(\sgn(\widehat \zeta) \neq \sgn(\zeta^*)) \leq P(\sgn(\zeta_{{\cA}_k}) \neq \sgn(\zeta^*_{{\cA}_k})) 
+ P( | ({\bz}^j)^T ({\by}^{k} - {\bZ}_{{\cA}_k} \bar{\zeta}_{{\cA}_k}) | > \lambda_n r_j, \exists~j \notin {\cal A}_k ).
$$

We now bound the two probabilities on the right hand side conditional on the set $\cal T$. For the brevity of abusing notations, we simply use $P(\cdot)$ to denote the conditional probability given $\cal T$ in the remaining of the proof. First, by (\ref{eqn:cond_sol1}),
\begin{eqnarray*}
& & P(\sgn(\bar \zeta_{{\cA}_k}) \neq \sgn(\zeta^*_{{\cA}_k})) \leq P \Big(|\zeta^*_j-\bar \zeta_j| \geq |\zeta^*_j|, \exists~j \in {\cal A}_k \Big ), \\
&\leq& P \Big ( \big| {\bf 1}_j^T ({\bZ}_{{\cA}_k}^T {\bZ}_{{\cA}_k})^{-1}  {\bZ}_{{\cA}_k}^T  {\bepsilon}_k \big | \geq \zeta^*_{min}/2 \Big ) + \\
& & P \Big (\big| {\bf 1}_j^T ({\bZ}_{{\cA}_k}^T {\bZ}_{{\cA}_k})^{-1}  {\bZ}_{{\cA}_k}^T {\bxi}_k \big | + \big| {\bf 1}_j^T ({\bZ}_{{\cA}_k}^T {\bZ}_{{\cA}_k})^{-1}  \lambda_n \bar{\bs}_{{\cA}_k} \big | \geq \zeta^*_{min}/2 \Big ),
\end{eqnarray*}
where ${\bf 1}_j$ is a vector of zeros except the $j$-th component being 1, and by (\ref{eqn:eigenbound}),
$$
\big| {\bf 1}_j^T ({\bZ}_{{\cA}_k}^T {\bZ}_{{\cA}_k})^{-1}  \lambda_n \bar{\bs}_{{\cA}_k} \big | \leq \big(c_1 \Lambda_{min}(d)\big)^{-1} \|n^{-1} \lambda_n \bar{\bs}_{{\cA}_k} \| \leq \big(c_1 \Lambda_{min}(d)\big)^{-1} n^{-1} \lambda_n d^{1/2} r_{max}({{\cA}_k}). 
$$
Furthermore, it follows from the definition of $\cal T$ and the initial Lasso estimates that 
\begin{eqnarray*}
& & \big| {\bf 1}_j^T ({\bZ}_{{\cA}_k}^T {\bZ}_{{\cA}_k})^{-1}  {\bZ}_{{\cA}_k}^T {\bxi}_k \big | \leq \| ({\bZ}_{{\cA}_k}^T {\bZ}_{{\cA}_k})^{-1}  {\bZ}_{{\cA}_k}^T {\bxi}_k \| \\
&\leq& \big(c_1 \Lambda_{min}(d)\big)^{-1} \|n^{-1} {\bZ}_{{\cA}_k}^T {\bxi}_k \| \leq \big (c_1 \Lambda_{min}(d)\big)^{-1} \big( \|n^{-1} {\bX}_{{\cA}_k^{\beta}}^T {\bxi}_k \| + \|n^{-1} \widehat{\by}_{{\cA}_k^{\omega}}^T {\bxi}_k \| \big ) \\
&\leq& (c_1 \Lambda_{min}(d))^{-1} \big ( O( n^{-1} d^{1/2} \lambda_{init})  + O(n^{-1/2} d^{1/2} (\log (p+q))^{1/2}) + O(n^{-2} d \lambda_{init}^2 K(d,d,3,\bM)) \big ).
\end{eqnarray*}
Since $\zeta^*_{min}/2$ is asymptotically larger than the upper bounds in the last two inequalities and $n^{-1}\| {\bf 1}_j^T (n^{-1} {\bZ}_{{\cA}_k}^T {\bZ}_{{\cA}_k})^{-1}  {\bZ}_{{\cA}_k}^T \| \leq n^{-1/2} (\Lambda_{min} (n^{-1} {\bZ}_{{\cA}_k}^T {\bZ}_{{\cA}_k}))^{-1/2} \leq n^{-1/2} \big( c_1 \Lambda_{min}(d) \big)^{-1/2}$, there exists some positive constant $c_3$ such that for sufficiently large $n$, 
\begin{eqnarray*}
P(\sgn(\bar \zeta_{{\cA}_k}) \neq \sgn(\zeta^*_{{\cA}_k})) &\leq& P \Big ( \big| {\bf 1}_j^T ({\bZ}_{{\cA}_k}^T {\bZ}_{{\cA}_k})^{-1}  {\bZ}_{{\cA}_k}^T {\bepsilon}_k \big | \geq \frac{1}{2} \zeta^*_{min},\exists~j \in {\cal A}_k \Big ) \\
&\leq& c_3 d \exp(-n c_1 \Lambda_{min}(d) (\zeta_{min}^*)^2/4 \tilde \sigma^*_{kk}).
\end{eqnarray*}

Next, to bound $P( | ({\bz}^j)^T ({\by}^{k} - {\bZ}_{{\cA}_k} \bar \zeta) | > \lambda_n r_j, \exists~j \notin {\cal A}_k )$, we have
\begin{eqnarray*}
& & P( | ({\bz}^j)^T ({\by}^{k} - {\bZ}_{{\cA}_k} \bar \zeta) | > \lambda_n r_j, \exists~j \notin {\cal A}_k ) \\
&=& P \Big ( | ({\bz}^j)^T ({\bH}_{{\cA}_k} ({\bxi}_k + {\bepsilon}_k) + {\bZ}_{{\cA}_k} ({\bZ}_{{\cA}_k}^T {\bZ}_{{\cA}_k})^{-1} \lambda_n \bar{\bs}_{{\cA}_k} ) |> \lambda_n r_j, \exists~j \notin {\cal A}_k  \Big )  \\ 
&\leq& P \Big ( |({\bz}^j)^T {\bH}_{{\cA}_k} ({\bxi}_k + {\bepsilon}_k)| + |({\bz}^j)^T {\bZ}_{{\cA}_k} ({\bZ}_{{\cA}_k}^T {\bZ}_{{\cA}_k})^{-1} \lambda_n \bar{\bs}_{{\cA}_k}| \geq \lambda_n r_j, \exists~j \notin {\cal A}_k \Big )，
\end{eqnarray*}
where $\bH_{{\cA}_k}=\bI - {\bZ}_{{\cA}_k} ({\bZ}_{{\cA}_k}^T {\bZ}_{{\cA}_k})^{-1} {\bZ}_{{\cA}_k}^T$. Note that (\ref{eqn:irrep}) implies that
$$
| ({\bz}^j)^T {\bZ}_{{\cA}_k}({\bZ}_{{\cA}_k}^T {\bZ}_{{\cA}_k})^{-1} \lambda_n \bar{\bs}_{{\cA}_k} | \leq \lambda_n r_j (1-\eta).
$$
for any $j \in {\cA}_k^c$. Therefore, 
$$
P( | ({\bz}^j)^T ({\by}^{k} - {\bZ}_{{\cA}_k} \zeta) | > \lambda_n r_j, \exists~j \notin {\cal A}_k ) \leq P \Big ( |({\bz}^j)^T {\bH}_{{\cA}_k} ({\bxi}_k + {\bepsilon}_k)| \geq \eta \lambda_n r_j, \exists~j \notin {\cal A}_k \Big ).
$$
By Lemma 11.3 of Zhou et al. (2009) and the fact that $\sigma^*_{jj}=1$, there exists a positive constant $c_4$ such that $P(\max_j n^{-1}({\bz}^j)^T {\bz}^j \geq c_4) \leq (p+q)^{-2}$. Conditional on the set $\{\max_j n^{-1}({\bz}^j)^T {\bz}^j \leq c_4\}$, $\|({\bz}^j)^T {\bH}_{{\cA}_k}\| \leq (c_4 n)^{1/2}$, $\|{\bxi}_k\| \leq O(n^{-1/2} d \lambda_{init})$ by Assumption (A1). Since $n^{-1/2} d \lambda_{init} =o(1)$, there exists some positive constant $c_5$ such that
$$
P( | ({\bz}^j)^T ({\by}^{k} - {\bZ}_{{\cA}_k} \zeta) | > \lambda_n r_j, \exists~j \notin {\cal A}_k ) \leq c_5 (p+q) \exp\left (- \frac{\eta^2 \lambda_n^2 r_{min}^2({\cal A}_k^c)}{2c_4n \tilde \sigma^*_{kk}} \right).
$$
Combining all the above results, for sufficiently large $n$, 
\begin{equation}
\begin{split}
P(\sgn(\widehat{\zeta}) & \neq \sgn({\zeta}^*)) \leq \\
& (p+q)^{-2} + c_3 d \exp\left (-\frac{n c_1 \Lambda_{min}(d)(\zeta_{min}^*)^2}{2 \tilde \sigma^*_{kk}} \right ) + c_5 (p+q) \exp\left (- \frac{\eta^2 \lambda_n^2 r_{min}^2({\cal A}_k^c)}{2c_4n \tilde \sigma^*_{kk}} \right ), \nonumber
\end{split}
\end{equation}
and the desired result follows immediately.


\noindent{\bf Proof of Theorem \ref{thm:est_cons}:} The solution of (\ref{eqn:cond_pls}) is the same as that of (\ref{eqn:cond_pls3}), where  $\widehat{\zeta}=(\widehat{\beta}_k^T,\widehat{\gamma}_k^T)^T$ and satisfies that 
$$
-2({\bz}^j)^T ({\by}^k-\bZ \widehat{\zeta})+\lambda_n r_j \sign(\widehat{\zeta}_j)=0,~\mbox{for any}~j \in {\cal A}_k.
$$
Let $\widehat{\bs}_{{\cal A}_k}=(r_j \sign(\widehat{\zeta}_j);~j \in {\cal A}_k)$, then $-2 {\bZ}_{{\cal A}_k}^T ({\by}^k-\bZ \widehat{\zeta})+\lambda_n \widehat{\bs}_{{\cal A}_k}=0$, or equivalently,
$$
\frac{1}{\sqrt{n}} {\bZ}_{{\cal A}_k}^T {\bZ}_{{\cal A}_k} (\widehat{\zeta}_{{\cal A}_k}-\zeta^*_{{\cal A}_k}) = \frac{1}{\sqrt{n}} {\bZ}_{{\cal A}_k}^T {\bepsilon}_k + \frac{1}{\sqrt{n}} {\bZ}_{{\cal A}_k}^T {\bxi}_k - \frac{\lambda_n}{2\sqrt{n}} \widehat{\bs}_{{\cal A}_k} - \frac{1}{\sqrt{n}} {\bZ}_{{\cal A}_k}^T {\bZ}_{{\cal A}_k^c} \widehat{\zeta}_{{\cal A}_k^c}.
$$

By the proof of Theorem \ref{thm:sel_cons}, on the set ${\cal T} \cap {\cal Y}_k$, $\Lambda_{min}(n^{-1} {\bZ}_{{\cal A}_k}^T {\bZ}_{{\cal A}_k}) \geq c_1 \Lambda_{min}(d)>0$. Let ${\bSigma}_k=n^{-1} {\bZ}_{{\cal A}_k}^T {\bZ}_{{\cal A}_k}$, on the set ${\cal T} \cap {\cal Y}_k$, for any $|{\cal A}_k| \times 1$ vector $\alpha$,
\begin{equation}
\begin{split}
\sqrt{n} s_k^{-1} {\alpha}^T (\widehat{\zeta}_{{\cal A}_k}-\zeta^*_{{\cal A}_k}) = \frac{1}{\sqrt{n}} s_k^{-1} {\alpha}^T {\bSigma}_k^{-1}  & {\bZ}_{{\cal A}_k}^T {\bepsilon}_k + \frac{1}{\sqrt{n}} s_k^{-1} {\alpha}^T {\bSigma}_k^{-1} {\bZ}_{{\cal A}_k}^T {\bxi}_k \\
& - \frac{\lambda_n}{2\sqrt{n}} s_k^{-1} {\alpha}^T {\bSigma}_k^{-1} \widehat{\bs}_{{\cal A}_k} - \frac{1}{\sqrt{n}} s_k^{-1} {\alpha}^T {\bSigma}_k^{-1} {\bZ}_{{\cal A}_k}^T {\bZ}_{{\cal A}_k^c} \widehat{\zeta}_{{\cal A}_k^c}. \nonumber
\end{split}
\end{equation}
We now show that on the set ${\cal T} \cap {\cal Y}_k$ the last three components converge to 0 in probability uniformly with respect to $\alpha$. First, the proof of Theorem \ref{thm:sel_cons} implies that $P(\widehat{\zeta}_{{\cal A}_k^c} = 0) \geq 1-(p+q)^{-2} \rightarrow 1$, and thus
$$
P \left ( \frac{1}{\sqrt{n}} s_k^{-1} {\alpha}^T {\bSigma}_k^{-1} {\bZ}_{{\cal A}_k}^T {\bZ}_{{\cal A}_k^c} \widehat{\zeta}_{{\cal A}_k^c} = 0 \right ) \longrightarrow 1.
$$
Second, by Assumption (A3) and the fact that $\| \alpha \|=1$,
\begin{eqnarray*}
\big | \frac{1}{\sqrt{n}} \lambda_n s_k^{-1} {\alpha}^T {\bSigma}_k^{-1} \widehat{\bs}_{{\cal A}_k} \big | &\leq& \frac{1}{\sqrt{n}} \lambda_n s_k^{-1} (\Lambda_{min}({\bSigma}_k))^{-1} \|\alpha\| \|\widehat{\bs}_{{\cal A}_k} \| \\
&\leq& \frac{1}{\sqrt{n}} \lambda_n s_k^{-1} (c_1 \Lambda_{min}(d))^{-1} d^{1/2} (\zeta^*_{min})^{-1} \longrightarrow 0,
\end{eqnarray*}
except on an event with probability tending to zero. Third, on the set ${\cal T} \cap {\cal Y}_k$,
\begin{eqnarray*}
\big | \frac{1}{\sqrt{n}} s_k^{-1} {\alpha}^T {\bSigma}_k^{-1} {\bZ}_{{\cal A}_k}^T {\bxi}_k \big | &\leq& s_k^{-1} \left \| \frac{1}{\sqrt{n}} {\alpha}^T {\bSigma}_k^{-1} {\bZ}_{{\cal A}_k}^T \right \|  \|{\bxi}_k\| = s_k^{-1} ({\alpha}^T {\bSigma}_k^{-1}{\alpha})^{1/2}  \|{\bxi}_k\| \\
&\leq& s_k^{-1} (c_1 \Lambda_{min}(d))^{-1/2}  \|{\bxi}_k\|  \longrightarrow 0,
\end{eqnarray*}
where $\|{\bxi}_k\| \leq O(n^{-1/2} d \lambda_{init})$ as in the proof of Theorem \ref{thm:sel_cons}. 

Therefore, we have on the set ${\cal T} \cap {\cal Y}_k$,
$$
\sqrt{n} s_k^{-1} {\alpha}^T (\widehat{\zeta}_{{\cal A}_k}-\zeta^*_{{\cal A}_k}) = \frac{1}{\sqrt{n}} s_k^{-1} {\alpha}^T {\bSigma}_k^{-1} {\bZ}_{{\cal A}_k}^T {\bepsilon}_k + o_p(1),
$$
where $\frac{1}{\sqrt{n}} s_k^{-1} {\alpha}^T {\bSigma}_k^{-1} {\bZ}_{{\cal A}_k}^T {\bepsilon}_k \stackrel{d}{\rightarrow} N(0,1)$ by verifying the conditions of the Linderburg-Feller central limit theorem as in Huang et al. (2008). Furthermore, on the set $({\cal T} \cap {\cal Y}_k)^c$, $|\sqrt{n} s_k^{-1} {\alpha}^T (\widehat{\zeta}_{{\cal A}_k}-\zeta^*_{{\cal A}_k})| \leq s_k^{-1} O_p( (p+q+d \lambda_n))$ by Theorem 1 of Huang et al. (2008), and $P(({\cal T} \cap {\cal Y}_k)^c) \leq (p+q)^{-2}$ by the proof of Theorem \ref{thm:sel_cons}. As $d \lambda_n = o(p+q)$, the desired asymptotic normality follows immediately.

\newpage

\begin{table}[!h]
\centering \caption{Averaged performance measures regarding $\widehat{\bB}$ and estimated
standard errors (in parenthesis) of aMCR, ALT and SEP,
over 50 replications. }
\bigskip
\begin{small}
\begin{tabular}{lccccccccc}
\hline
& & \multicolumn{3}{c}{Estimation Accuracy} & & \multicolumn{4}{c}{Selection Accuracy} \\
\cline{3-5} \cline {7-10} 
 & & $\|{\bDelta}_B\|_F$ & $\|{\bDelta}_B\|_1$ & $\|{\bDelta}_B\|_{\infty}$ &  &  Dist & Spe & Sen & Mcc \\
\hline
& & \multicolumn{8}{c}{\it Model 1: $(p, q,n) = (100, 100, 250)$}  \\
SEP & &  37.7(.41) & 2.84(.055) & 2.86(.103) & &  .01(.000) & .99(.000) & .55(.005) & .64(.004) \\
ALT & &  $-$ & $-$ & $-$ & &  $-$ & $-$ & $-$ & $-$ \\
aMCR & &  18.6(.30) & 2.01(.040) & 2.19(.055) & &  .01(.000) & 1.0(.000) & .55(.005) & .68(.004) \\
& &  \multicolumn{8}{c}{\it Model 2: $(p, q,n) = (50, 50, 250)$}  \\
SEP & &  19.0(.261) & 2.75(.062) & 2.63(.048) & &  .03(.001) & .96(.001) & .63(.007) & .58(.005) \\
ALT & &  13.2(.764) & 2.49(.063) & 2.53(.062) & &  .07(.003) & .86(.006) & .77(.016) & .46(.009) \\
aMCR & &  10.3(.260) & 2.04(.062) & 2.22(.048) & &  .02(.001) & .99(.000) & .60(.006) & .69(.005) \\
& &  \multicolumn{8}{c}{\it Model 3: $(p, q,n) = (10, 25, 250)$}  \\
SEP & &  4.16(.073) & 2.64(.053) & 1.45(.036) & &  .09(.003) & .84(.006) & .76(.008) & .60(.007) \\
ALT & &  3.94(.099) & 2.64(.059) & 1.48(.041) & &  .11(.003) & .79(.020) & .81(.010) & .59(.014) \\
aMCR & &  3.28(.074) & 2.18(.058) & 1.31(.039) & &  .07(.002) & .96(.003) & .69(.009) & .71(.007) \\
& &  \multicolumn{8}{c}{\it Model 4: $(p, q,n) = (200, 1000, 250)$}  \\
SEP & & 447.2(1.69) & 10.37(.121) & 4.32(.250) & & .01(.000) & 1.0(.000) & .56(.002) & .63(.001) \\
ALT & &  $-$ & $-$ & $-$ & &  $-$ & $-$ & $-$ & $-$ \\
aMCR & & 235.3(.96) & 7.06(.090) & 3.32(.078) & & .01(.000) & 1.0(.000) & .54(.001) & .66(.001) \\
& &  \multicolumn{8}{c}{\it Model 5: $(p, q,n) = (200, 800, 250)$}  \\
SEP & & 355.1(1.53) & 8.83(.104) & 4.05(.075) & & .01(.000) & 1.0(.000) & .55(.001) & .63(.001) \\
ALT & &  $-$ & $-$ & $-$ & &  $-$ & $-$ & $-$ & $-$ \\
aMCR & & 186.4(.86) & 6.19(.090) & 3.28(.063) & & .01(.000) & 1.0(.000) & .54(.001) & .66(.001) \\
& &  \multicolumn{8}{c}{\it Model 6: $(p, q,n) = (200, 400, 150)$}  \\ 
SEP & &  177.6(.96) & 5.34(.070) & 4.15(.213) & &  .01(.000) & 1.0(.000) & .56(.002) & .63(.002) \\
ALT & &  $-$ & $-$ & $-$ & &  $-$ & $-$ & $-$ & $-$ \\
aMCR & &  93.6(.71) & 3.80(.063) & 3.01(.057) & & .01(.000) & 1.0(.000) & .55(.002) & .66(.001) \\
\hline
\end{tabular}
\end{small}
\label{tab:simB}
\end{table}

\begin{table}[!h]
\centering \caption{Averaged performance measures regarding $\widehat{\bOmega}$ and estimated
standard errors (in parenthesis) of aMCR, ALT and SEP,
over 50 replications. }
\bigskip
\begin{small}
\begin{tabular}{lcccc}
\hline
& \multicolumn{4}{c}{Selection Accuracy} \\
\cline{2-5}
  & Dist & Spe & Sen & Mcc \\
\hline
& \multicolumn{4}{c}{\it Model 1: $(p, q,n) = (100, 100, 250)$}  \\
SEP & .09(.001) & .82(.001) & .77(.006) & .31(.003) \\
ALT & $-$ & $-$ & $-$ & $-$ \\
aMCR & .02(.000) & .99(.000) & .47(.007) & .61(.005) \\
& \multicolumn{4}{c}{\it Model 2: $(p, q,n) = (50, 50, 250)$}  \\
SEP & .09(.001) & .82(.002) & .77(.006) & .41(.004) \\
ALT & .05(.009) & .93(.021) & .52(.028) & .53(.016) \\
aMCR & .05(.001) & .99(.000) & .50(.007) & .64(.005) \\
& \multicolumn{4}{c}{\it Model 3: $(p, q,n) = (10, 25, 250)$}  \\
SEP & .09(.003) & .83(.004) & .77(.010) & .53(.007) \\
ALT & .11(.013) & .83(.038) & .57(.038) & .47(.023) \\
aMCR & .08(.022) & .99(.012) & .54(.009) & .65(.007) \\
& \multicolumn{4}{c}{\it Model 4: $(p, q,n) = (200, 1000, 250)$}  \\
SEP & .07(.000) & .86(.000) & .73(.001) & .12(.000) \\
ALT & $-$ & $-$ & $-$ & $-$  \\
aMCR & .03(.000) & 1.0(.000) & .38(.002) & .52(.001) \\
& \multicolumn{4}{c}{\it Model 5: $(p, q,n) = (200, 800, 250)$}  \\
SEP & .08(.000) & .85(.000) & .73(.002) & .13(.000) \\
ALT & $-$ & $-$ & $-$ & $-$   \\
aMCR & .00(.000) & 1.0(.000) & .39(.002) & .53(.002) \\
& \multicolumn{4}{c}{\it Model 6: $(p, q,n) = (200, 400, 150)$}  \\
SEP & .08(.000) & .83(.000) & .75(.002) & .17(.000) \\
ALT & $-$ & $-$ & $-$ & $-$  \\
aMCR & .01(.000) & 1.0(.000) & .41(.003) & .55(.002) \\ 
\hline
\end{tabular}
\end{small}
\label{tab:simO}
\end{table}

\begin{table}[!h]
\centering \caption{Averaged predictive square errors, numbers of selected genes and their estimated standard errors over 50 replications. }
\bigskip
\begin{small}
\begin{tabular}{lcc}
\hline
  & Pse & Num.gene \\
\hline
SEP & 1.21(.011) & 74.9(2.22) \\
ALT & $-$ & $-$ \\
aMCR & 1.19(.012) & 65.2(1.75) \\
\hline
\end{tabular}
\end{small}
\label{tab:gene}
\end{table}

\end{document}